\documentclass[runningheads]{llncs}

\usepackage[T1]{fontenc}
\usepackage{graphicx}
\usepackage{fontawesome}
\usepackage{marvosym}
\usepackage{booktabs}
\usepackage{array}
\newcolumntype{L}{>{\RaggedRight\arraybackslash}X}
\usepackage{float}

\usepackage{rotating}
\usepackage{array}
\usepackage{multicol}
\usepackage{multirow}
\usepackage{xcolor}

\usepackage{amsmath} 
\usepackage{makecell} 

\usepackage{adjustbox} 

\usepackage{tabularx,ragged2e,booktabs}

\definecolor{customgreen}{HTML}{44ac64}
\definecolor{customred}{HTML}{ff0000}

\usepackage{cite}

\usepackage{orcidlink}
\usepackage{xurl}
\usepackage{hyperref}

\begin{document}

\title{Automated Cardiac Adipose Tissue Segmentation in Computed Tomography: A Literature Review}
%
\titlerunning{Review: Cardiac Adipose Tissue Segmentation Methods}
%
\author{Andreas W. Aspe\inst{1}\orcidlink{0009-0003-2798-6586} \and Jonas Jalili Pedersen\inst{2}\orcidlink{0000-0002-9496-7655} \and Andreas Ohrt Johansen\inst{2}\orcidlink{0000-0002-7391-9945} \and Klaus Fuglsang Kofoed\inst{2}\orcidlink{0000-0001-9742-1554} \and Kristine Aavild Sørensen\inst{1,3}\orcidlink{0000-0003-1434-697X} \and Rasmus R. Paulsen\inst{1}\orcidlink{0000-0003-0647-3215} \and Josefine Vilsbøll Sundgaard\inst{1,3}\orcidlink{0000-0003-2872-4660}}
\authorrunning{F. Author et al.}
%
\institute{DTU Compute, Technical University of Denmark, Kongens Lyngby, Denmark \email{awias@dtu.dk} \and Heart Centre, Rigshospitalet, Copenhagen, Denmark, Denmark \and Novo Nordisk A/S, Søborg, Denmark}
\maketitle                

\begin{abstract}
This review provides an overview of recent advancements in automated segmentation methods on Computed Tomography (CT) for two types of cardiac fat: Epicardial adipose Tissue (EAT) and Pericardial Adipose Tissue (PAT). These fat deposits, separated by the pericardium, have been linked to various cardiovascular diseases, with EAT receiving the most research attention. Their complex anatomical context makes manual quantification highly time-consuming and prone to considerable inter-observer variability. Automated methods effectively address these complications, offering a more efficient and consistent solution. This study encompasses a broad range of methods, spanning AI as well as non-AI approaches. Additionally, it presents the remaining challenges, including the need for larger annotated public datasets and optimized attenuation thresholds for contrast-enhanced CT. It is demonstrated that automated methods are able to achieve segmentation results comparable to the quality of human annotation, proving their potential as a clinical tool for discovering new biomarkers and enhancing patient outcomes.

\end{abstract}

\section{Introduction}

Cardiovascular Diseases (CVDs) represent the leading cause of mortality and disability, accounting for more than 42.5\% of all annual deaths in the European Region~\cite{CVD1_web}. Major CVDs include ischemic heart disease, stroke, and cardiac arrhythmogenic disorders~\cite{Mensah2023}. Adiposity and obesity play a central role in cardiovascular health and disease by influencing systemic metabolism, vascular function, and inflammatory processes~\cite{Oikonomou2019}. Adiposity is commonly assessed using Body Mass Index (BMI), a widely used but simplistic measure that does not account for fat distribution in the body. Adipose tissue may be further categorized according to anatomical location, typically separated into subcutaneous adipose tissue underneath the skin and visceral adipose tissue surrounding the inner organs. Recent advances in medical imaging allow for quantification of visceral adipose tissue volumes. Among these, cardiac adipose tissue has been a matter of growing interest since pioneering research described the anatomy and clinical relationship with the heart early in this century~\cite{Iacobellis2005}. Adipose tissue located underneath the pericardium, Epicardial Adipose Tissue (EAT), has been associated with cardiac diseases including heart failure, ischemic heart disease, and atrial arrhythmogenic disorder~\cite{Iacobellis2022,Konwerski2022}. Although potentially less clinically prominent, Pericardial Adipose Tissue (PAT), situated above the pericardium, has also been associated with atrial fibrillation and inflammation~\cite{Greif2013}.

\begin{figure}[t]
\includegraphics[width=\textwidth]{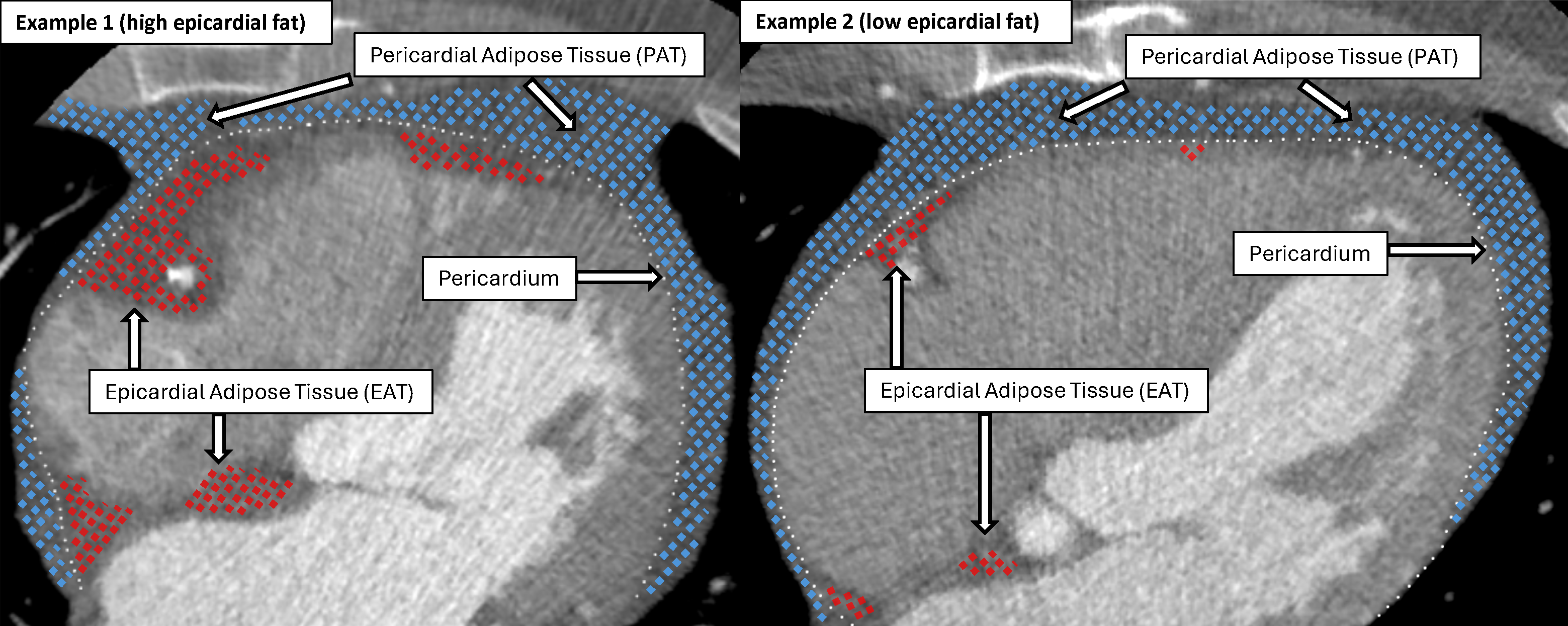}
\caption{The anatomy of EAT, PAT and pericardium on CCTA~\cite{Fuchs2023}.} \label{fig:fat_CT}
\end{figure}

Computed tomography (CT) has been established as an important clinical tool in the assessment of cardiac disease~\cite{Vrints2024}. Modern CT is standardized in clinical practice as a time-efficient image acquisition method, delivering low radiation doses. Contrast-enhanced Cardiac CT Angiography (CCTA) is an effective imaging modality for cardiac segmentation, offering superior visualization of cardiac structures such as the pericardium, also compared to other image modalities such as MRI due to higher resolution~\cite{Militello2019,Li2021}. Fig.~\ref{fig:fat_CT} shows CCTA scans from two patients in the Copenhagen General Population Study (CGPS)~\cite{Fuchs2023} with EAT and PAT highlighted in red and blue, respectively. Today, segmenting EAT and PAT from CT largely relies on labor-intensive visual qualitative and semi-quantitative assessments with notable limitations, including observer variability~\cite{Barbosa2011,Greco2022}.

Recently, Artificial Intelligence (AI) has enabled automated analysis of EAT and PAT, decreasing manual labor while showing competitive performance compared to manual annotations~\cite{Bencevic2022,Greco2022}. These methods facilitate large-scale studies and clinical correlation analyses. For example, Kroll et al.~\cite{Kroll2021} utilized a 3D U-Net to segment EAT and PAT in just 5–10 seconds per patient, investigating its association with the Coronary Artery Calcium Score in a large cohort.

The purpose of this paper is to give a comprehensive technical overview of
recent image analysis methods for cardiac adipose tissue segmentation using CT. It is a highly active research field with many recent publications. The latest reviews are from 2022~\cite{Bencevic2022,Greco2022} and many contributions have been made since, which appeals for an updated review. Additionally, there is a need to establish a clear and consistent terminology for the different types of cardiac adipose tissues. Methods within the last 10 years are covered and were found through citation chasing and search databases such as PubMed and Google Scholar, applying numerous relevant keywords. Only methods developed for CT and methods distinguishing between EAT and PAT are considered.

\section{Cardiac adipose tissue and anatomy}
\begin{figure}[b]
    \centering
    \includegraphics[width=\linewidth, trim=0cm 0cm 0cm 0.7cm, clip]{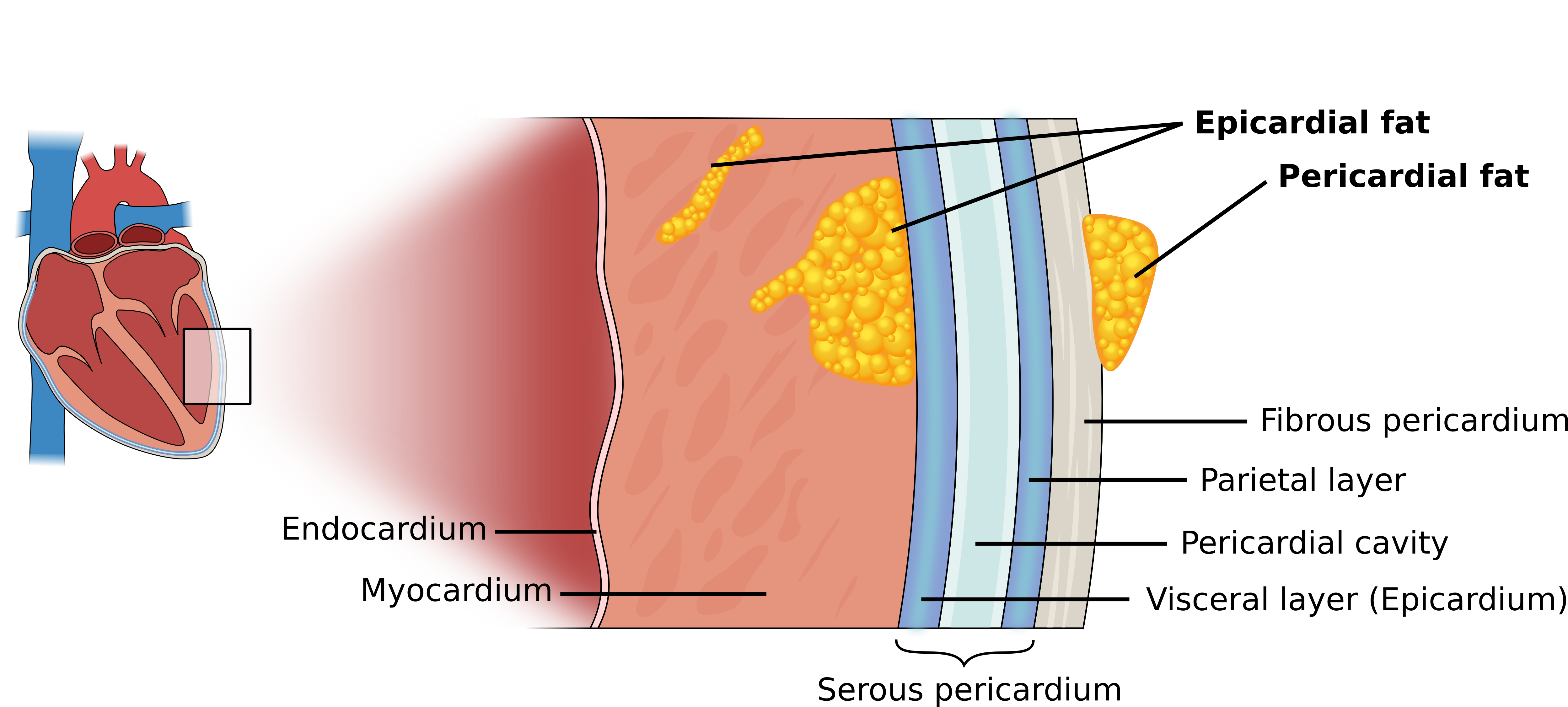}
    \caption{The pericardium and location of cardiac adipose tissues.}
    \label{fig:fat_sketch}
\end{figure}
Cardiac adipose tissue can be divided into segments, anatomically separated by the pericardial sac (the pericardium), a thin protective layer around the heart. The pericardium has two main parts: the fibrous pericardium and the serous pericardium. Additionally, the serous pericardium has two layers: the outer parietal layer which is attached to the fibrous pericardium, and the inner visceral layer, also known as the epicardium, which lies directly on the myocardium (the heart muscle)~\cite{Sacks2007}. There is a small fluid-filled space between the parietal and the visceral layers called the pericardial cavity. Fat located underneath the epicardium is termed Epicardial Adipose Tissue (EAT), whereas fat located around the heart, outside the epicardium, is called Pericardial Adipose Tissue (PAT). The anatomy of the pericardium and cardiac adipose tissues is illustrated in Fig.~\ref{fig:fat_sketch}. The distinction between EAT and PAT is important, as they have different anatomic relationships and different biomolecular properties~\cite{Iacobellis2005}.

Adipose tissue on CT is characterized by a distinct unique Hounsfield Unit (HU) range, usually defined as (-190,~-30)~HU~\cite{Yoshizumi1999}. This attenuation range is substantially lower than other tissues, such as muscles, myocardium and blood vessels, making adipose tissue easily identifiable through thresholding.

There is terminological disagreement in the medical literature regarding the definition of the adipose tissues surrounding the outside of the heart. Some literature defines pericardial adipose tissue to be the union of fat located inside and outside the pericardium~\cite{Bencevic2022}. Other studies refer to the fat outside the pericardium as paracardial adipose tissue and define pericardial adipose tissue as fat specifically located inside the pericardium itself and within the pericardial cavity~\cite{Greco2022,Konwerski2022}. However, since the pericardium is a very thin structure, the layers cannot be distinguished on CT and the pericardium is typically only slightly visible due to a small increase in attenuation of the fibrous layer. In terms of reviewing previous papers, we will acknowledge any wording or misconceptions and adjust the descriptions to fit the terminology defined in this paper.
\section{Methods in literature}
An overview of the most important methods and results for EAT segmentation can be found in Tab.~\ref{tab:EAT}. The 'Dataset' column notes the total number of CT volumes used to train and test each segmentation model. The label '(CE)' indicates exclusive use of contrast-enhanced scans, no label implies non-contrast scans, while '(Both)' specifies the use of both non-contrast and contrast-enhanced scans. The star (*) indicates if a given method is based on the same publicly available dataset by VisualLab~\cite{VisualLab}. While the vast majority of methods focus on EAT, some studies are also concerned with PAT. These are listed in Tab.~\ref{tab:PAT}.

There are generally two approaches for doing EAT and PAT segmentation on CT: 1)~direct segmentation of adipose tissue or 2)~pericardium delineation followed by thresholding. Direct segmentation is the conventional approach, where a model generates a mask or probability map directly for the target. However, due to the irregular distribution of cardiac adipose tissue, it may be beneficial to instead segment the pericardium and apply thresholding to the inner and outer regions, in which EAT and PAT are defined, using the HU range of fat.

The earliest approaches for automated EAT segmentation emerged in the 2000s, using traditional image analysis methods based on intensities, textures, and region growing algorithms~\cite{Bandekar2006,Dey2008_QFAT}. Subsequently, atlas-based methods gained prominence, while also beginning to differentiate between EAT and PAT~\cite{Shahzad2013,Ding2014}. This paradigm shifted as machine learning gained popularity~\cite{Rodrigues2015a} and soon after, deep learning took over. Nowadays, most of the proposed work uses some kind of deep learning, while a few methods still rely on more traditional non-AI techniques.

\begin{table}[htbp]
\fontsize{10}{12}\selectfont  
\caption{Overview of methods for EAT segmentation. CE = contrast-enhanced scans, Both = non-contrast and contrast-enhanced scans, $\rho_c$ = Lin concordance correlation, $\rho_s$ = Spearman correlation, * = Visual Lab dataset~\cite{VisualLab}.}
\label{tab:EAT}
\begin{tabularx}{\textwidth}{@{}lllLll@{}}
        \toprule
        \textbf{Authors} & \textbf{Year} & \textbf{Dim.} & \textbf{Method} & \textbf{Dataset} & \textbf{Results} \\
        
        \midrule
        \multicolumn{6}{c}{Deep Learning} \\
        \midrule
        Commandeur et al. \cite{Commandeur2018} & 2018 & 2D & Two CNNs & 250 & DSC = 82.3\% \\
        Commandeur et al. \cite{Commandeur2019} & 2019 & 2.5D & CNN & 850 & DSC = 87.3\% \\
        Zhang et al. \cite{Zhang2020} & 2020 & 2D & Two U-Nets & 20* & DSC = 91.2\% \\ 
        He et al. \cite{He2020} & 2020 & 3D & U-Net & 200 (CE) & DSC = 88.7\% \\
        Li et al. \cite{Li2021} & 2021 & 2.5D & U-Net & 103 (CE) & DSC = 93.3\% \\
        Siriapisith et al. \cite{Siriapisith2021} & 2021 & 3D & U-Net & 260 (Both) & DSC = 90.1\% \\ 
        Molnar et al. \cite{Molnar2021} & 2021 & 3D & CNN & 1811 & DSC = 90\% \\
        Ben\v{c}evi\'{c} et al. \cite{Bencevic2021a} & 2021 & 2D & U-Nets, Polar transform & 20* & DSC = 78.4\% \\
        Ben\v{c}evi\'{c} et al. \cite{Bencevic2021b} & 2021 & 2.5D & U-Net & 20* & DSC = 85.7\% \\
        Li et al. \cite{Li2022} & 2022 & 3D & U-Net & 70 (Both) & DSC = 92.6\% \\ 
        Qu et al. \cite{Qu2022} & 2022 & 2D & CNN, U-Net & 103 & DSC = 88.3\% \\
        Hoori et al. \cite{Hoori2022} & 2022 & 2.5D & DeepLab-v3-plus + bisect & 89 & DSC = 88.5\% \\ 
        Bartoli et al. \cite{Bartoli2022} & 2022 & 2D & U-Net + xHRNet & 115 & DSC = 85\% \\
        West et al. \cite{West2023} & 2023 & 3D & U-Net & 3720 (CE) & $\rho_{c}=0.972$ \\
        Qu et al. \cite{Qu2023} & 2023 & 3D & CNN, Swin UNETR & 724 & DSC = 83.9\% \\
        Silva et al. \cite{SantosdaSilva2024} & 2024 & 2D & pix2pix & 20* & DSC = 98.7\% \\ 
        Miller et al. \cite{Miller2024} & 2024 & 2.5D & Conv. LSTM & 1138 & $\rho_{s}=0.91$ \\
        Tang et al. \cite{Tang2024} & 2024 & 2D & U-Net & 108 (CE) & DSC = 89.6\% \\
        Kuo et al. \cite{Kuo2024} & 2024 & 3D & U-Net & 157 (CE) & DSC = 87.0\% \\
        Wang et al. \cite{Wang2025} & 2025 & 2.5D & U-Net & 50 (CE) & DSC = 93.9\% \\

        \midrule
        \multicolumn{6}{c}{Machine Learning} \\
        \midrule
        Rodrigues et al. \cite{Rodrigues2016} & 2016 & 2D & Atlas, RF & 20* & DSC = 98.1\% \\
        Norlén et al. \cite{Norlen2016} & 2016 & 3D & Atlas, RF, graph cuts & 30 (CE) & DSC = 91\% \\
        Priya and Sudha \cite{Priya_Sudha2019} & 2019 & 2D & GLCM, region growing & 20* & DSC = 98.7\% \\ 
        Kazemi et al. \cite{Kazemi2020} & 2020 & 2D & GLCM, RF & 20* & DSC = 97.8\% \\   

        \midrule
        \multicolumn{6}{c}{Non-AI methods} \\
        \midrule
        Zlokolica et al. \cite{Zlokolica2017} & 2017 & 2D & Fuzzy c-means, ellipse fitting & 10 & DSC = 69\% \\ 
        Militello et al. \cite{Militello2019} & 2019 & 2D & Manual ROIs, interpolation & 145 (Both) & DSC = 93.1\% \\ 
        Rebelo et al. \cite{Rebelo2022} & 2022 & 2D & Template matching, Convex Hull & 20* & DSC = 77.3\% \\ 
        \bottomrule
\end{tabularx} 

\end{table}

\begin{table}[htbp]
\fontsize{10}{12}\selectfont  
\caption{Overview of methods for PAT segmentation. CE = contrast-enhanced scans, * = Visual Lab dataset~\cite{VisualLab}.}
\label{tab:PAT}
\begin{tabularx}{\textwidth}{@{}lllLll@{}}
        \toprule
        \textbf{Authors} & \textbf{Year} & \textbf{Dim.} & \textbf{Method} & \textbf{Dataset} & \textbf{Results} \\
        
        \midrule
        \multicolumn{6}{c}{Deep Learning} \\
        \midrule
        Commandeur et al. \cite{Commandeur2018} & 2018 & 2D & Two CNNs & 250 & \multicolumn{1}{c}{-} \\
        He et al. \cite{He2021} & 2020 & 3D & U-Net & 422 (CE) & DSC $\geq 0.85$ \% \\
        Hoori et al. \cite{Hoori2023} & 2023 & 2.5D & DeepLab-v3-plus + bisect & 89 & DSC = 82.5\% \\
        Silva et al. \cite{SantosdaSilva2024} & 2024 & 2D & pix2pix & 20* & DSC = 98.4\% \\

        \midrule
        \multicolumn{6}{c}{Machine Learning} \\
        \midrule
        Rodrigues et al. \cite{Rodrigues2016} & 2016 & 2D & Atlas, RF & 20* & \multicolumn{1}{c}{-} \\
        Priya and Sudha \cite{Priya_Sudha2019} & 2019 & 2D & GLCM, region growing & 20* & DSC = 98.3\% \\ 
        Kazemi et al. \cite{Kazemi2020} & 2020 & 2D & GLCM, RF & 20* & DSC = 98.4\% \\   

        \bottomrule
\end{tabularx} 
\end{table}

\subsection{Non-AI Methods}\label{sec:nonAI}
Zlokolica et al.~\cite{Zlokolica2017} propose a slice-based semi-automatic method for EAT segmentation. The method is based on morphological operations to extract the heart, followed by a fuzzy c-means clustering algorithm and geometric ellipse fitting. The method requires the user to place a seed manually once for each patient. The authors only present the Dice Similarity Coefficient (DSC) of 8 patients, showing considerable variation. The mean is presented in Tab. \ref{tab:EAT}.

Militello et al.~\cite{Militello2019} propose a technically simple semi-automatic method to assist manual annotation, making it faster and more precise. The authors developed a GUI, requesting the user to manually draw pericardium delineations on a self-chosen number of slices. Linear interpolation is then performed to fill out the delineation gaps, with the option for users to refine the curves and, when satisfied, define the adipose tissue threshold range for extracting EAT. The process is 4-8 times faster than slice-based manual segmentation. 

Rebelo et al.~\cite{Rebelo2022} isolate the heart region by using Otsu's Method, connected component analysis, and template matching. The pericardium contour is refined by computing the convex hull and using morphological operations, allowing the authors to access EAT by thresholding. The method is implemented in an open-source software called HARTA.

\subsection{Machine Learning Methods}\label{sec:machinelearning}
Both the method by Rodrigues et al.~\cite{Rodrigues2016} and Norlén et al.~\cite{Norlen2016} utilize a multi-atlas approach to do registration, followed by training a Random Forest (RF) classifier for tissue classification. Whereas Rodrigues et al. classify EAT and PAT directly, Norlén et al. use Random Forest to estimate if a voxel is just inside, just at the boundary, or just outside of the pericardium or neither of the three. Segmentation is completed using graph cuts from a Markov random field.

Kazemi et al.~\cite{Kazemi2020} enhance CT slices using Contrast Limited Adaptive Histogram Equalization (CLAHE) and extract features such as contrast, correlation, energy, etc. by using the gray-level co-occurrence matrix (GLCM) and Gabor filters. A Random Forest classifier is trained on these features to directly segment four classes, among them being EAT and PAT. Priya and Sudha~\cite{Priya_Sudha2019} also utilize GLCM for feature extraction, which serves as input to a shallow Grey Wolf-Optimized (GWO) neural network for classifying fat or non-fat slices. A region-growing algorithm combined with Fruitfly optimization is then applied to segment EAT, PAT, and the pericardium.

\subsection{Deep Learning Methods}\label{sec:deeplearning}

\subsubsection{Network types}
Most networks in this review are based on U-Nets. A few methods~\cite{Kuo2024,Zhang2020,Bencevic2021a,Bencevic2021b} use the conventional U-Nets architecture as originally presented in Ref.~\cite{Unet}, while others add slight modifications such as batch normalization or residual connections~\cite{Tang2024,West2023,Li2022}. Some methods present more complex U-Net variants incorporating concepts like attention and transformers~\cite{He2020,Siriapisith2021,Qu2023,Wang2025}. Molnar et al.~\cite{Molnar2021} utilize a Convolutional Neural Network (CNN) with an architecture very similar to a U-Net, however, instead of downsampling, a raw CT volume path is provided as input across multiple layers at different resolutions.

Other networks than the U-Net are also proposed. Miller et al.~\cite{Miller2024} employ a convolutional Long-Short Term Memory model (LSTM). Hoori et al.~\cite{Hoori2022,Hoori2023} apply DeepLab-v3-plus which has ResNet-18 as a backbone. Bartoli et al.~\cite{Bartoli2022} propose the network called xHR-NET, which maintains High-Resolution representations throughout the network. Lastly, Silva et al.~\cite{SantosdaSilva2024} apply pix2pix, a generative conditional adversarial network primarily designed for image-to-image translation tasks.

\subsubsection{Multi-step appproaches}
A common approach for solving the segmentation task is to first determine whether a slice or path contains the heart or not~\cite{Bartoli2022,Qu2022,Qu2023,Commandeur2018,Commandeur2019}. Commandeur et al.~\cite{Commandeur2018} use two sequential CNN's for EAT and PAT segmentation, where the first network is responsible for slice selection. Upsampling of hidden layers is then used as input to the second network, responsible for the delineation of the pericardium. Combining the outputs, the segmentation masks are obtained by thresholding. A year later, the same authors propose a simplified method in which both tasks are handled by a single condensed CNN. Similarly, Qu et al.~\cite{Qu2022} use a CNN to do slice classification and a variant of the U-Net for the actual segmentation of EAT; in this case direct. Bartoli et al.~\cite{Bartoli2022} implement a classification head in the center of a U-Net to identify slices containing the heart, before performing the actual segmentation with the network xHR-Net.

\subsubsection{Attention}
Two methods utilize 3D U-Nets with incorporated attention gates to focus on EAT boundary structures.~\cite{He2020,Siriapisith2021}. He et al. adopt the same attention-based 3D U-Net in Ref.~\cite{He2021} as in Ref.~\cite{He2020} to do EAT and PAT segmentation, respectively. Qu et al.~\cite{Qu2022} present a multi-scale residual attention U-Net, and Miller et al.~\cite{Miller2024} incorporate attention head into the LSTM for doing pericardium segmentation. Although Miller et al. do not report DSC, they note a promising Spearman volume correlation of 0.91 compared to expert manual segmentations.

\subsubsection{Transformer}
The first transformer-based architecture is presented by Qu et al.~\cite{Qu2023} in 2023. As in their earlier work~\cite{Qu2022}, a 2D CNN selects candidate slices for EAT to reduce the consumption of GPU memory. For the segmentation, the authors employ a 3D transformer model Swin UNETR for its effectiveness in modeling global dependencies. Recently, Wang et al.~\cite{Wang2025} introduce Convolutional Transformer modules into the classical U-Net, yielding state-of-art results; however, the method was tested on only 10 patients.

\subsubsection{Transfer Learning}
Transfer learning is used by two methods to accommodate the shortcoming of few data samples. Siriapisith et al.~\cite{Siriapisith2021} train on 220 CT scans and fine-tune on a smaller dataset of 40 CCTA scans, achieving a DSC of 90.1\%, but a slightly lower number, 88.2\%, is reported for CCTA. This last result closely matches the performance reported by He et al.~\cite{He2020}, who use a similar network trained exclusively on CCTA data, with both studies utilizing datasets of comparable total size. Hoori et al.~\cite{Hoori2022} perform segmentation with the network DeepLab-v3-plus, which is pre-trained on the ImageNet dataset.

\subsubsection{2.5D methods}
Some methods~\cite{Commandeur2019,Hoori2022,Li2021,Miller2024,Bencevic2021b,Wang2025} balance spatial precision with computational efficiency by employing multi-slice approaches, i.e. multiple slices are inputted to segment a single target slice. Ben\v{c}evi\'{c} et al.~\cite{Bencevic2021b} input the CT slice, alongside an additional image with every pixel representing the slice depth, to a U-Net, arguing that the pericardial region strongly depends on slice depth. Hoori et al.~\cite{Hoori2022} give three consecutive CT slices as input, segmenting the middle one. They also apply 'bisection', which entails splitting the heart into two halves at its widest point. This ensures the network processes slices with increasing pericardial curvature, simplifying learning through a consistent image progression. The same method is applied by Hoori et al. to segment PAT in Ref.~\cite{Hoori2023}.
\section{Discussion}

\subsection{Data}
Many previous methods rely on very small datasets, making it difficult to assess the model's ability to generalize. While training set size is important, the size of the test set is equally critical. For example, Siriapisith et al.~\cite{Siriapisith2021} use 200 non-contrast scans for training but only 20 scans for testing. Although this dataset is larger than in many other studies, the test set most likely does not capture a representative variation in anatomy. Contrary, Miller et al.~\cite{Miller2024} utilize a much larger test set comprising 638 patients from four different sites, increasing the likelihood that the test set better reflects a real-life distribution.

To avoid overfitting, it is crucial to create a test set without data leakage. Data leakage might happen if slices from the same patient are present in both the training and test set as the method adapts to a certain anatomy. Silva et al.~\cite{SantosdaSilva2024} conduct random slice-based testing rather than patient-based testing, which likely explains why it achieves a significantly higher DSC compared to other methods. A similar approach may have been used in methods presenting similar high results~\cite{Rodrigues2016,Kazemi2020,Priya_Sudha2019}, as their data splitting is not thoroughly explained.

A major limitation in the field of cardiac fat segmentation is the lack of publicly available datasets with manual annotations. The only open source dataset with labeled EAT and PAT volumes is provided by Visual Lab \cite{VisualLab}, created by Rodrigues et al.~\cite{Rodrigues2015a}. This scarcity has led numerous methods to rely on and train their models using this dataset~\cite{Bencevic2021a,Bencevic2021b,Kazemi2020,Priya_Sudha2019,Rebelo2022,Rodrigues2016,SantosdaSilva2024,Zhang2020}. However, it is very small, comprising only 20 non-contrast CT scans, which is insufficient for developing robust, generalizable models and for proper validation. High performance on this dataset may be a result of overfitting. Additionally, the annotations lack clarity: true labels are represented with an RGB (Red, Green and Blue) color, but 31\% of voxels exhibit overlapping color codes. This ambiguity affects EAT and PAT quantification across models, as no standardized method for differentiation in such cases has been defined.

It is evident that more open-source data is needed. It is of great interest to have large common datasets of high quality that can be used for benchmarking. Without this, it will be difficult to compare the quality of different models. Also, there is currently no publicly available dataset on contrast-enhanced scans. This would be of high interest, as European Guidelines have implemented CCTA as a recommended imaging tool for detecting heart disease in symptomatic individuals~\cite{Vrints2024}, and also the pericardium is more visible on CCTA~\cite{Kuo2024}, potentially easing the job for data-driven segmentation models.

\subsection{Cardiac fat attenuation range}
Many of the reviewed methods rely on the characteristic attenuation values of adipose tissue, especially when segmenting adipose tissue by pericardium delineation and thresholding, making the choice of an accurate interval crucial. While this interval is typically defined as (-190,~-30)~HU~\cite{Yoshizumi1999}, the exact threshold values are subject to debate, as it varies with e.g. acquisition parameters and the use of contrast medium~\cite{Marwan2019,Yin2022,Xu2018}. This variability is also reflected in methods using non-contrast scans, where proposed HU intervals include (-200,~-30)~\cite{Rebelo2022,SantosdaSilva2024,Rodrigues2016,Kazemi2020,Bencevic2021a,Bencevic2021b}, (-210,~-10)~\cite{Qu2022}, and (-140,~-30)~\cite{Bartoli2022}, with the most methods adhering to the standard interval (-190,~-30).

\begin{figure}[t]
\begin{minipage}{0.55\textwidth}
\includegraphics[width=1.0\linewidth]{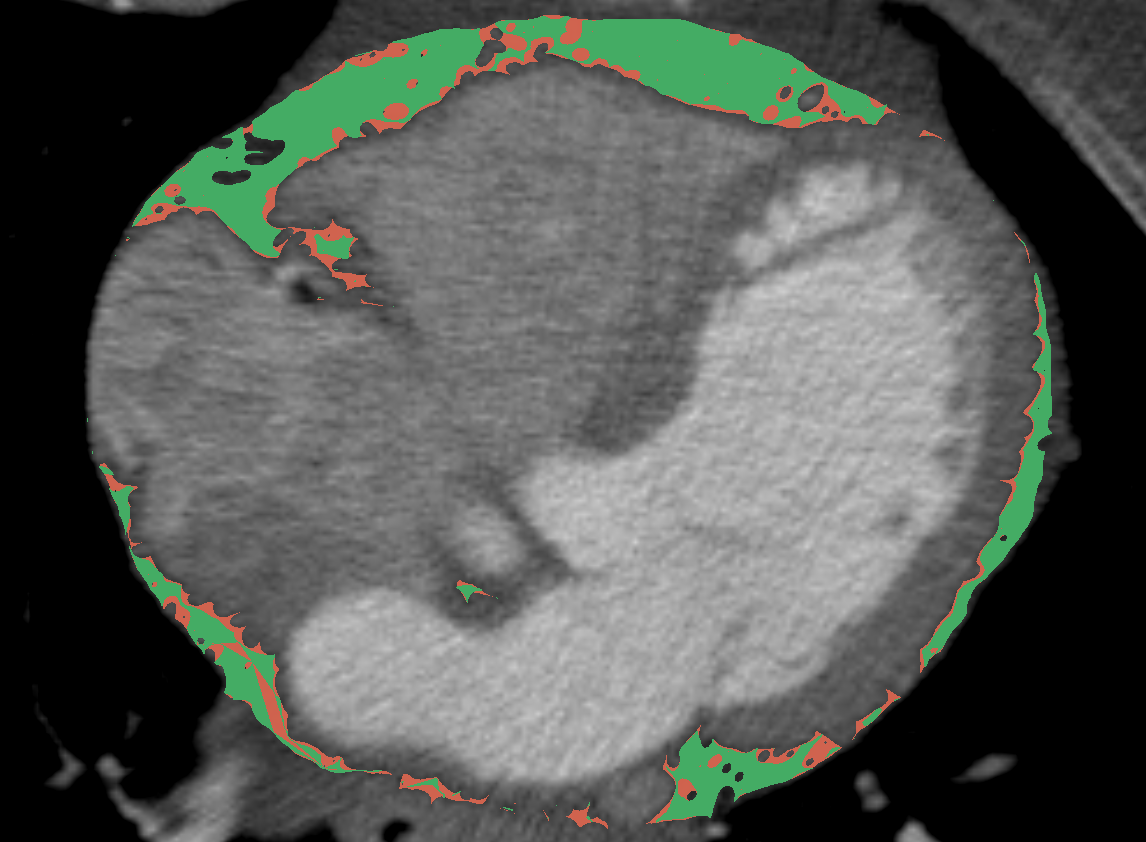}
\end{minipage}
\begin{minipage}{0.2\textwidth}
 \begin{tabular}{ll}
     \hspace{1mm} \textcolor{customgreen}{\textbf{Green}} & \vspace{1mm} \\
     \hspace{1mm} HU attenuation range: $[-190,-30]$ & \\
     \hspace{1mm} Volume: $4.37 \ \text{cm}^3$ \vspace{3mm} & \\
     \hspace{1mm} \textcolor{orange}{\textbf{Orange}} & \vspace{1mm} \\
     \hspace{1mm} HU attenuation range: $[-190,0]$ & \\
     \hspace{1mm} Volume: $5.33 \ \text{cm}^3$ ($\sim 20\%$ increase) & \\
 \end{tabular}
\end{minipage}
\caption{The difference between EAT attenuation thresholds on a CCTA slice~\cite{Fuchs2023}.}\label{fig:HU_tresholds}
\end{figure}

While variations in the lower threshold appear to have minimal impact, the upper threshold is particularly critical for contrast-enhanced CT. Studies report that using a standard fixed upper threshold of -30 significantly underestimates EAT in contrast-enhanced scans compared to non-contrast CT~\cite{Marwan2019,Xu2018}. For contrast-enhanced scans, studies suggest that an upper threshold closer to 0 HU is more accurate~\cite{Marwan2019,Xu2018}. Fig. \ref{fig:HU_tresholds} illustrates this difference using a contrast-enhanced CT-slice from the CGPS dataset\cite{Fuchs2023}, annotated locally by a medical doctor. EAT regions are marked for two upper thresholds: -30 HU (\textcolor{customgreen}{green}) and 0 HU (\textcolor{orange}{orange}), with -190 HU as the lower limit in both cases. Increasing the upper threshold from -30 HU to 0 HU resulted in an approximate 20\% increase in estimated EAT volume.

For methods using contrast-enhanced CT scans, the proposed intervals are (-190,~-30)~HU\cite{Wang2025,Li2022,He2020,Tang2024,Kuo2024}, (-192,~-30)~HU~\cite{Norlen2016} and (-175,~-15) HU \cite{Li2021}. These intervals fail to account for the appropriate upper threshold range, resulting in systematic under-segmentation of EAT. The last interval, while closer to the ideal range, still likely underestimates the total EAT volume.

\subsection{Model performance and challenges}
Multiple studies report better segmentation performance at slices in the middle of the heart, with poorer performance near the upper and lower end~\cite{Commandeur2018,Rebelo2022,Tang2024}. This is to be expected, as these regions are also more difficult to segment manually. At the lower end, the pericardium is typically thinner and may therefore be harder to identify. The anatomy becomes more intricate at the upper end, where the pericardium connects to the aorta, pulmonary veins, and arteries, introducing ambiguity about its exact boundary. To truly compare the performance of the methods, a clear definition of the heart limits should be defined. For instance, West et al.~\cite{West2023} define the limits to be the bifurcation of the pulmonary trunk superiorly and the apex of the pericardium inferiorly, whereas Bartoli et al.~\cite{Bartoli2022} set the limits to be the top of the left atrium and the last slice in which the left ventricle was identified. A consensus is needed to compare models and give a robust measure of the performances in the extremities.

Variation in anatomy is a major issue for automated methods in this field, as EAT and PAT exhibit significant individual differences in distribution and composition. Moreover, several methods report reduced performance in cases with high EAT and PAT~\cite{Commandeur2018,Miller2024}, likely due to anatomical variations in patients with higher BMI. Similarly, Kazemi et al.~\cite{Kazemi2020} report poor performance when attempting to segment the pericardial border due to anatomical variability.

Cardiac adipose tissue lacks a neat, uniform structure. Sometimes it appears as clusters, other times as fragments of smaller chunks. For this reason, most methods in this review segment EAT by delineating the pericardium rather than performing direct segmentation. The pericardium provides a relatively smooth and consistent outline, making it easier to identify. However, with this approach, segmentation inaccuracies tend to occur near the pericardial borders, and since the majority of both EAT and PAT reside close to the pericardial sac, even small misalignments can significantly impact segmentation performance.

In general, many methods demonstrate a performance competitive with manual segmentation. Strong correlations with manual estimations of EAT and PAT volume and attenuation are reported. Molnar et al.~\cite{Molnar2021} concluded 99.4\% of the automated segmentations to be successful. However, the last few unsuccessful scans did not have any easily identifiable characteristics. While automated approaches show some variability, manual segmentation does as well, to a comparable extent. Bartoli et al.~\cite{Bartoli2022} accounts for an inter-observer DSC of $\sim3\%$, whereas others report an inter-observer variation in volume of 10.4\%~\cite{Barbosa2011}. 


In the past, processing time was a major limiting factor for segmentation methods, but advancements in computational power and model efficiency have largely resolved this issue. While Rodrigues et al.~\cite{Rodrigues2016} reported processing times of several hours, recent methods perform interference in a few seconds\cite{Qu2022,Miller2024,Tang2024}, enabling real-time clinical use.

\subsection{Future directions and clinical perspectives}
With the growing number of precise and efficient deep-learning algorithms for the segmentation of cardiac adipose tissue, more research should be put into identifying novel biomarkers and uncovering new clinical outcomes. Currently, most work is focused on total EAT volume and/or attenuation, however, not many are focusing on the effects of possible regional accumulations, making this an intriguing area for future exploration.

An evident next step is to broaden the scope of segmentation models to include other types of cardiac fat. Clinically, there is a great interest in adipose tissue located around the coronary arteries~\cite{Simantiris2024}, a subset of EAT. However, automated methods directly addressing this have yet to emerge.
\section{Conclusion}
Developing accurate methods to measure EAT and PAT is vital for enhancing CVD risk assessment and understanding their role in cardiovascular health and disease progression. This review demonstrates a broad set of methods for automated segmentation, including deep learning, machine learning and non-AI methods. Several of these are highly accurate with low processing times, highlighting the potential of automated models based on medical image analysis and AI to support clinicians and contribute to medical diagnostics.
\begin{credits}
\subsubsection{\ackname} This study was supported by a research grant from Novo Nordisk A/S.

\subsubsection{\discintname}
The authors have no competing interests to declare that are relevant to the content of this article.

\end{credits}

\bibliographystyle{splncs04}
\bibliography{bibliography}

\end{document}